# Reinforcement learning in large, structured action spaces: A simulation study of decision support for spinal cord injury rehabilitation

Nathan Phelps[1,*], Stephanie Marrocco[2,3], Stephanie Cornell[2], Dalton L. Wolfe[2,3], and Daniel J. Lizotte[1,4]

[*]Corresponding author

[1]Department of Computer Science, University of Western Ontario, London, Ontario, Canada

[2]Gray Centre for Mobility and Activity, Parkwood Institute, Lawson Health Research Institute, London, Ontario, Canada

[3]Health and Rehabilitation Sciences, University of Western Ontario, London, Ontario, Canada

[4]Department of Epidemiology & Biostatistics, University of Western Ontario, London, Ontario, Canada

*Email: nphelps3@uwo.ca; Postal Address: 1151 Richmond St., London, ON, N6A 3K7

**Abstract:** Reinforcement learning (RL) has helped improve decision-making in several applications. However, applying traditional RL is challenging in some applications, such as rehabilitation of people with a spinal cord injury (SCI). Among other factors, using RL in this domain is difficult because there are many possible treatments (i.e., large action space) and few patients (i.e., limited training data). Treatments for SCIs have natural groupings, so we propose two approaches to grouping treatments so that an RL agent can learn effectively from limited data. One relies on domain knowledge of SCI rehabilitation and the other learns similarities among treatments using an embedding technique. We then use Fitted Q Iteration to train an agent that learns optimal treatments. Through a simulation study designed to reflect the properties of SCI rehabilitation, we find that both methods can help improve the treatment decisions of physiotherapists, but the approach based on domain knowledge offers better performance. Our findings provide a "proof of concept" that RL can be used to help improve the treatment of those with an SCI and indicates that continued efforts to gather data and apply RL to this domain are worthwhile.

**Competing Interests:** The authors have no competing interests to declare.

**Acknowledgements:** We thank Melissa Fielding, Deena Lala, Patrick Stapleton, Bonnie Chapman, Heather Askes, Rozhan Momen, and the rest of the physiotherapists from the spinal cord injury and acquired brain injury rehabilitation programs at Parkwood Institute for their input in the development of the simulator.

**Funding:** This work was supported by St. Joseph's Health Care London and the Natural Sciences and Engineering Research Council of Canada (NSERC) through an Alexander Graham Bell Canada Graduate Scholarship.

# 1. Introduction

Spinal cord injury (SCI) is characterized by damage and resulting dysfunction to the motor, sensory, and/or autonomic nervous systems associated with trauma or disease processes leading to traumatic or non-traumatic SCI, respectively. The functional consequences can therefore be wide-ranging across these systems, with varying degrees of muscle paralysis, sensory impairment, and autonomic dysfunction such as problems with cardiovascular control, thermoregulation, or bowel, bladder, or sexual function [1], [2]. In general, the more rostral (higher) the damage to the spinal cord, the more body systems that will be affected. With respect to motor function, persons with damage to the cervical (neck) area of the spinal cord will have impairments to both lower and upper limb muscles and are diagnosed as having tetraplegia, while persons with damage to the thoracic (back) or lumbar (lower back) area of the spinal cord will have impairments to the muscles of the thorax and/or the lower limbs only and are diagnosed as having paraplegia. Given the functional consequences of SCI are dependent on both the severity and level of the damage to the nervous system, in addition to a variety of other factors such as pre-morbid condition, additional secondary complications, and psychosocial influences, there is a significant degree of heterogeneity in the presentation of persons with SCI [1], [2].

The prevalence of SCI, including both traumatic and non-traumatic SCI, was estimated in Canada for 2010 to be 85 556 persons – 37 313 (44%) people living with tetraplegia and 48 243 (56%) people living with paraplegia [3]. Total health care costs for persons with SCI have been estimated to be almost seven and a half times higher than non-SCI individuals, with an observed net cost of $336 100 CAD from the perspective of the Ontario Ministry of Health and Long-Term Care [4].

Following a traumatic SCI in Canada, persons often spend a short amount of time in acute care (13 – 44 days) [5], followed by a stay at a rehabilitation facility. Length of stays in Canada are generally determined by injury severity, functional status, and other contributing factors (e.g., secondary complications, discharge location) [6]. Rehabilitation stays in Canada are typically between 50 – 124 days for traumatic SCI [5]. During an inpatient stay, patients are typically seen daily by a physiotherapist. Access to specialized outpatient services (e.g., within a rehabilitation hospital) is dependent on where the patient lives.

To measure a patient's status, it is important to gather measurable outcomes using standardized assessments. A standard assessment approach across Canadian rehabilitation settings involves the Standing and Walking Assessment Tool (SWAT) [7], [8]. This assessment tool involves characterizing the patient's function related to the person's ability to sit, stand, and walk, and the relative level of assistance required across 12 discrete categories of function. In rehabilitation practice, these are referred to as stages with the general idea that persons will progress from one stage to the next as they regain more function.

Best rehabilitative care for persons with SCI is deemed to involve treatment by an interdisciplinary team of health care professionals in specialized, integrated centres with care delivered as early as possible following onset with discharge to community services characterized by ongoing outpatient care and follow-up [9]. Physiotherapists are key members of this team, and they typically focus on therapies aimed at enhancing strength or endurance with goals targeting functional abilities reflected in the SWAT stages, such as transferring from wheelchair to bed, standing, walking, or upper extremity gross or fine motor control [10]. Specific patient goals are selected in collaboration with the patient by considering patient preferences as well as information gathered from integrated and comprehensive assessments. Often, patient goals that are particularly relevant to physiotherapists include enhancing trunk control, standing balance, or walking function. Physiotherapists will deliver a combination of therapy activities, which are individualized to the patient to help achieve their goals. Optimally choosing therapy activities is challenging for a variety of reasons. These include the limited time available during a rehabilitation stay, given the myriad of issues that need to be addressed to support a person's safe return to their home or other discharge destination. Additionally, there are many possible therapeutic activities that could be considered for a given patient, yet little clear evidence to guide the therapist as to which combination of therapies might be most appropriate for that individual. For example, recent approaches to categorize different therapeutic activities that may be considered by physiotherapists in SCI rehabilitation have described multi-layered systems comprising at least three or four hierarchies of therapeutic practice resulting in potentially hundreds of options, especially when one considers variations in intensity or other adaptations [11], [12]. In a recent scoping review, Marrocco et al. identified at least eighteen different classification systems that have been developed to try to characterize the various therapies that might be conducted by therapists, although there has been no consensus or consistency with respect to the application of these systems in practice (personal communication). This lack of consensus reflects the complex nature of the field and the varying contexts under which therapies might be delivered.

Decision support can be used to help physiotherapists choose which treatments to use, leading to improved patient outcomes. A common approach for developing decision support systems is reinforcement learning (RL), which has been used in several health care applications, including treatment of chronic conditions such as cancer (e.g., [13]–[15]), diabetes (e.g., [16]–[18]), anemia (e.g., [19]–[21]), HIV (e.g., [22]–[24]), and mental illnesses (e.g., [25]–[27]), as well as critical care (e.g., [28]–[30]). See [31] for a more comprehensive coverage of previous uses of RL in health care. Despite the multitude of applications of RL in health care, to our knowledge, RL has yet to be applied to SCI rehabilitation. This may be due to the challenges associated with applying RL to this domain. From an RL perspective, the problem can be characterized in the following way:

- Limited training data is available (i.e., training data is expensive)
- The action space (number of possible actions) is very large
- Actions are selected simultaneously (multi-action selection)
- The training data is obtained in advance through the decisions of an intelligent agent (the physiotherapists), but the policy followed is unknown
- The actions are structured in a manner such that they can be grouped

The limited availability of training data, large action space, and selection of multiple actions simultaneously make using RL challenging. However, the structure within the actions can be used to help mitigate the negative effects of these attributes. In this study, we focused primarily on addressing the large action space and limited training data. We leave it to further studies to develop methods that explicitly consider the multi-action component of this problem.

The methods we developed in this study are designed to take advantage of the *structure* of the treatments in SCI rehabilitation, meaning prior information about which treatments are expected to perform similarly in similar contexts. We propose two methods that place the treatments into groups, allowing the agent to learn about a treatment from data about related treatments. The main contributions of this work are 1) new methodology for RL in large action spaces with prior information and 2) a simulator for such data that mimics the properties of data obtained from SCI rehabilitation. Section 2 provides a background of RL, representation learning, and SCI rehabilitation at Parkwood Institute; Section 3 describes the simulator we developed to imitate the rehabilitation process and the methods used to learn a policy from this data; Section 4 outlines our results; Section 5 discusses limitations of the study and next steps; and Section 6 concludes the paper.

## 2. Background

### 2.1 Reinforcement learning

Reinforcement learning (RL) aims to build decision-making systems that base their decisions on the current and past state of the world, the previous actions undertaken, and the actions that may be taken in the future [32]. Along with supervised learning (e.g., regression, classification) and unsupervised learning (e.g., clustering), it is a subfield within the field of machine learning. In RL, the goal is to train a learner/decision-maker, known as the *agent*. In the SCI rehabilitation setting, the physiotherapist can be thought of as an agent. We want to train the agent to interact with its *environment* in an "optimal" fashion (where "optimal" is defined using a numerical reward system). In an RL framework, the environment is everything outside of the agent with which the agent interacts. The current situation in the environment is described by its *state*, which for our problem could include factors such as the patient's abilities or age. The agent interacts with its environment by selecting an *action*, which in our case refers to a therapeutic activity, or treatment. After selecting an action, the agent receives from the environment both a *reward* and information about the new state. The reward is a numeric representation of the desirability of an outcome, so standardized assessments like the SWAT can be used here. A function of the rewards, known as the *return*, can be constructed to place less value on rewards obtained further along in the future. The expected value of this return is the value that the agent attempts to maximize with its actions. Figure 1 illustrates the general framework of an RL problem.

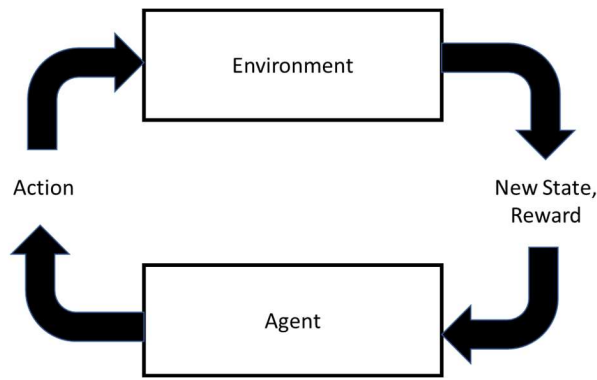

Figure 1: The general framework of a reinforcement learning (RL) problem.

Markov decision processes (MDPs) are mathematical formalisms used for modelling RL problems. Consider discrete time points, $t = 0, 1, 2, \ldots$ At each time $t$, the agent receives information about the state, $S_t$, from the set of all possible states, $S$, and chooses an action, $A_t$, from the set of all possible actions, $A$. It then receives a reward, $R_t$, and information about the environment's new state, $S_{t+1}$. The transition to $S_{t+1}$ occurs based on $P$, a function that defines the probability of transitioning to $S_{t+1}$ given $S_t$ and $A_t$. For each time step $t$, the data collected takes the following form: $(S_t, A_t, R_t, S_{t+1})$. This process can either continue indefinitely or until an end state is reached. In cases where an end state is reached, the process is known as an *episode*, the data of which consists of a *trajectory* $(s_1, a_1, r_1, s_2, a_2, r_2, \ldots, s_T, a_T, r_T)$ and an associated return given by $\sum_{t=1}^{T} \gamma^{t-1} r_t$, where $\gamma$ is a discount factor between zero and one. The discount factor reflects that rewards obtained further into the future may be less valuable than rewards obtained sooner.

The agent chooses its actions based on the current state according to a *policy*, which is a mapping from states to actions. Through experience, the goal is to learn a policy that maximizes the expected return, called an *optimal policy*. Finding an optimal policy is not as simple as choosing the action that maximizes the expected immediate reward; some actions may not yield a large immediate reward, but instead cause a transition to a new state with a large expected reward – or cause a transition to a new state that will eventually lead to another state with a large expected reward. For this reason, it is necessary to have some way of assessing the value of being in each state in $S$. A *state value function* is a function that estimates the expected return achieved by following a given policy starting from a given state. Alternatively, a *state-action value function* can be used. For each state-action pair $(s, a)$ in $S \times A$, the state-action value function estimates the expected return from taking action $a$ in state $s$ and following a given policy thereafter.

## 2.2 Representation learning

In statistical modelling tasks, it is common to transform the input variables (also known as features or predictors) in some way to improve the model. Representation learning, or feature learning, is a set of techniques used to generate new features by transforming the original input. Autoencoding, principal component analysis, word embedding, and clustering are all forms of representation learning (e.g., [33]). Our work uses word embedding and clustering techniques for representation learning.

Word embeddings are a way of representing words in vector space in such a way that similar words (i.e., words that occur in similar contexts) should be close to each other. GloVe [34] is a popular model used to generate word embeddings using a corpus of training data. It makes use of a co-occurrence matrix, which is a representation of the number of times words occur together within a context window (i.e., a number of words considered to the left and right of the context word).

Clustering is an unsupervised learning approach used to identify similar observations and place them into groups. To do this, a similarity metric (e.g., Euclidean distance) must be used to measure the similarity between observations. This choice of metric can potentially have a substantial impact on the resulting clusters, as can the choice of clustering algorithm. There are several different clustering algorithms, including single linkage, max linkage, and K-means [35]. As discussed by Ben-David [36], each of these algorithms emphasizes different requirements, so the choice of clustering algorithm must be made with the application in mind. For example, K-means clustering prioritizes that each group is composed of a similar number of observations, while the linkage algorithms do not.

**2.3 Spinal cord injury rehabilitation at Parkwood Institute**

At Parkwood Institute in London, Ontario, patients sustaining an SCI are admitted for rehabilitative care from across the geographic region of Southwestern Ontario, which represents a catchment area of over 1.5 million people. In the few years prior to the beginning of this study, data had been collected about the rehabilitation of approximately 90 patients. For each patient, the data included weekly documentation of their SWAT stage, as well as the treatments done each week. Although the data provided important information, physiotherapists were able to encode treatments however they liked, meaning the data included various ways of documenting the same treatment, personal acronyms, spelling errors, etc. At the time of writing, there are ongoing efforts to make the collection and storage of SCI rehabilitation data at Parkwood Institute broader and more uniform.

# 3. Methods

This study developed a simulator for SCI rehabilitation and new methodology to use RL to learn from this data. Both were implemented in R [37].

**3.1 A simulator for physical rehabilitation of spinal cord injury**

To facilitate the development of RL methods for SCI treatment selection, we created a simulator to imitate the rehabilitation process for a patient. This simulator was constructed from a combination of existing SCI data (described in Section 2.3) and extensive input from physiotherapists from the Research 2 Practice team at Parkwood Institute. Although existing data were used to guide parameter choices for the simulator, the focus when creating the simulator was placed on the properties of SCI rehabilitation. Thus, while the simulator generally reflects the characteristics of SCI rehabilitation (e.g., many possible treatments, treatments chosen simultaneously, transition probability cannot approach one), it is not a comprehensive representation of the dynamics of SCI rehabilitation.

There are two main benefits of using a simulator in this study. First, although it is anticipated that SCI treatment data will accrue through the Research 2 Practice team's project, the amount of data available at the time of this study was limited, so much so that it may not be reasonable to expect to be able to reliably train an agent from the data using any methodology, and therefore not feasible to evaluate the potential of different training methodologies using the data alone. Second, the use of a simulator facilitates evaluation of the methods in a way that would not be possible with real data; the same set of simulated patients can be treated using different treatment selection processes and the results from each process can be compared to one another.

The simulator was designed such that each patient falls into one of 12 stages ranging from zero, the most impaired, to 11, a terminal stage indicating unimpaired mobility. These stages were chosen to represent the 12 stages of the SWAT. The simulator has 11 groups composed of 10 treatments each, resulting in 110 total treatments. A treatment plan of eight treatments was chosen at each time point, representing the treatments a physiotherapist chooses each week for their patients. These treatments were chosen according to a simulated physiotherapist's perceived treatment benefits, which are numeric representations of how beneficial the physiotherapist thinks that treatment will be in that situation. The patient's probability of improving to the next SWAT stage is then based on the actual treatment benefits associated with the selected treatments. Further details about the simulator can be found in Appendix 1.

## 3.2 Learning dynamic treatment regimes

Applying RL to treatment selection for SCI rehabilitation is particularly challenging because of limited available data and because many treatments are selected simultaneously. Consequently, several foundational RL approaches which implicitly assume a large amount of data and a small action space are ineffective. In Appendix 2, we provide details of other RL approaches that we applied to this problem but that were ineffective, as well as the reason(s) why they did not work.

A key challenge in this setting is high variance in state-action value estimates due to a potentially very large action space combined with limited training data. As in regularized regression like lasso regression [38] and ridge regression [39], it can be beneficial to use a model with more bias to reduce the variance of the parameter estimates, in this case by reducing the number of parameters in the models. With this in mind, we propose grouping the actions since it is known that there is some structure to the actions in SCI rehabilitation. With an infinite amount of data, models using grouped data would be less effective than the traditional approach because of their relatively high bias. However, the reduced variance in the models with grouped actions should allow the agent to learn faster, facilitating its ability to learn something meaningful from the limited amount of data available.

We present two approaches to grouping actions that make use of prior knowledge that the action space is "structured" in that there is an a priori belief that some actions are likely to perform well in the same stage. The first approach requires an explicit mapping of actions to groups. In other words, the domain knowledge must be so extensive that the structure of the actions is completely known. The second approach requires less domain knowledge; under the assumption that actions are selected intelligently, the action selections in the data can provide guidance with regards to the grouping of the actions. Similar actions (i.e., actions that should belong to the same group) should be used in similar situations. In this case, explicit domain knowledge regarding the grouping of actions is not needed.

### 3.2.1 Domain knowledge based grouping

Using their knowledge of SCI rehabilitation, physiotherapists can group treatments together based on characteristics such as orientation (lying, sitting, standing, etc.), level of movement (static vs. dynamic), purpose (priming, impairment management, etc.), and level of independence (assisted vs. independent). To reflect this idea in the simulator, the actions were placed into 11 groups of 10 actions. We represented the physiotherapists grouping SCI treatments using their knowledge by using the known 11 groups of actions. For the remainder of this paper, we will refer to the agent that groups actions in this way as the domain knowledge based grouping (DKBG) agent.

### 3.2.2 Treatment embedding based grouping

Both time and domain expertise are needed to formulate an action grouping system as described in Section 3.2.1. It is possible that it may be infeasible to create such a system, or may not be worth the investment if a simpler alternative approach can achieve similar results. For this reason, we suggest an alternative approach that groups the actions using representation learning. Since the training data is generated through an intelligent behaviour policy (the physiotherapists' decisions), we can assume that the treatments that are commonly selected simultaneously are similar to one another.

Inspired by word embedding, we propose mapping each of the individual treatments to a point in vector space (i.e., treatment embedding). The training data obtained from SCI rehabilitation is structured in a similar way to a corpus used to generate word vectors; each weekly treatment plan is analogous to a text document and the treatments themselves are analogous to the words. Using the package text2vec [40], we generated treatment vectors by implementing the GloVe [34] modeling framework using 50 epochs and an *x_max* argument of 10. To create the co-occurrence matrix needed to train the model, we used an infinite context window since the ordering of treatments within the same treatment plan is arbitrary. An example of some of the co-occurrence pairs generated from a treatment plan are shown in Table 1. The bolded treatment is the context treatment and the treatments within the context window (all other treatments) are underlined. Note that a treatment plan is a set, so the order of the treatments is unimportant.

| Example Treatment Plan | Co-occurrence Pairs |
|---|---|
| **24**, 26, 33, 34, 38, 39, 42, 50 | (24, 26); (24, 33); (24, 34); (24, 38); (24, 39); (24, 42); (24, 50) |
| 24, **26**, 33, 34, 38, 39, 42, 50 | (26, 24); (26, 33); (26, 34); (26, 38); (26, 39); (26, 42); (26, 50) |
| 24, 26, **33**, 34, 38, 39, 42, 50 | (33, 24); (33, 26); (33, 34); (33, 38); (33, 39); (33, 42); (33, 50) |
| 24, 26, 33, **34**, 38, 39, 42, 50 | (34, 24); (34, 26); (34, 33); (34, 38); (34, 39); (34, 42); (34, 50) |

Table 1: An illustration of some of the co-occurrence pairs resulting from an example weekly treatment plan. The context treatment is bolded and treatments within the context window are underlined.

After each treatment is represented as a vector, the treatments can be grouped using a clustering algorithm. In the simulator, each of the action groups is composed of 10 treatments. Since K-means clustering is sensitive to imbalance in the number of treatments in each group, we chose to use this approach. The treatments were grouped into 11 groups (the same number of groups as created using domain knowledge) using the kmeans function with the number of initial configurations and the maximum number of iterations both set to 1000. For the remainder of this paper, we will refer to the agent that groups actions in this way as the treatment embedding based grouping (TEBG) agent.

### 3.2.3 Defining state-action pairs

To compute estimates of state-action values, we must first define state-action pairs. When actions are selected simultaneously, the action space grows very quickly if all combinations of actions are considered independently. One simple way of dealing with this challenge is to assume that the expected outcome following any action choice is unrelated to any other action choices made at the same timepoint (i.e., choosing not to control for impacts of other actions). To be clear, this assumption is not plausible, but represents the simplest possible way of handling the multi-action component of the problem. The original data had eight actions in each row of data. Based on the aforementioned assumption, each row of data was mapped to eight rows, so that each action was represented by a single row and the stage, number of weeks remaining, reward, and next stage were the same for all eight rows. We tested this approach using an illustrative example and found that it ranked the treatments in the same way as a more standard approach; Appendix 3 provides further details about this comparison. To facilitate an agent learning from the grouped data, one further alteration to the data was necessary. For each action, its action group was stored in the data instead of the individual action. An example is shown in Table 2.

Original Data:

| Stage | No. of Weeks Remaining | Actions | Reward | Next Stage |
|---|---|---|---|---|
| 3 | 2 | 24, 26, 33, 34, 38, 39, 42, 50 | 0 | 3 |

Mapping of Action to Action Group:

| Action | Action Group |
|---|---|
| 24 | 3 |
| 26 | 3 |
| 33 | 4 |
| 34 | 4 |
| 38 | 4 |
| 39 | 4 |
| 42 | 5 |
| 50 | 5 |

Adjusted Data:

| Stage | No. of Weeks Remaining | Action Group | Reward | Next Stage |
|---|---|---|---|---|
| 3 | 2 | 3 | 0 | 3 |
| 3 | 2 | 3 | 0 | 3 |
| 3 | 2 | 4 | 0 | 3 |
| 3 | 2 | 4 | 0 | 3 |
| 3 | 2 | 4 | 0 | 3 |
| 3 | 2 | 4 | 0 | 3 |
| 3 | 2 | 5 | 0 | 3 |
| 3 | 2 | 5 | 0 | 3 |

Table 2: Table 2a (top) shows a possible row from the simulated data in its original format, Table 2b (middle) shows the mapping of each applicable action to its action group, and Table 2c (bottom) shows the transformation of the original row into eight rows in the adjusted dataset using the grouping of actions.

### 3.2.4 Defining the reward function and return

We opted to give a reward of zero for each step until the final week of treatment. At this point, the reward is the final stage reached by the patient. We used a discount factor of one (i.e., undiscounted), so the return is simply the reward given at the end of the treatment period. This setup was chosen to reflect the goal of maximizing a patient's end-of-treatment abilities.

### 3.2.5 Estimating the optimal state-action value function

Since our training data is obtained in advance through following a less than optimal policy, we used an off-policy batch learning algorithm, Fitted Q Iteration [33], to estimate the optimal state-action value function. Fitted Q Iteration operates on *tuples* of data of the form $(s_t, a_t, r_t)$. The tuples provided to the algorithm are those in the adjusted data (Table 2c). Our requirement for convergence was that none of the coefficients of the following regression model changed by more than 0.0001 from one iteration to the next:

$$Q(s, a_g, No.of\ Weeks\ Remaining)$$
$$= \beta_0 + \beta_1(No.of\ Weeks\ Remaining)$$
$$+ \sum_{actionGroup} \beta_{actionGroup} I(actionGroup = a_g)$$
$$+ \sum_{stage} \beta_{stage} I(stage = s) \quad (1)$$
$$+ \sum_{actionGroup,stage} \beta_{actionGroup,stage} I(actionGroup = a_g) I(stage = s)$$

### 3.2.6 Assessing agent performance

To assess the agents, we simulated the rehabilitation process for 1000 patients, choosing their treatments using various selection processes. To facilitate meaningful evaluation of the agents' decision-making and their effectiveness in improving treatment selection for people with an SCI, we selected treatments using the four policy definitions below:

1. $\pi_{PT}$: Treatments were selected according to their perceived treatment benefit. This treatment selection process was identical to the treatment selection process used to generate the training data and was intended to reflect the approach taken by physiotherapists.
2. $\pi_{Agent}$: Since the state-action values learned by the model are based on groups of treatments rather than individual treatments, the state-action values do not provide a way of directly choosing individual treatments. To use the learned group-level state-action values, we needed to decide how they should be used to rank and choose individual treatments. We opted to multiply the group state-action value by the proportion of selections for each treatment in the given stage relative to the number of selections for its group in the stage. An alternative approach would be to uniformly sample from the treatments in the best group, but this would essentially reduce the problem to choosing the best group of treatments and we do not believe that this is a challenge for physiotherapists to do without decision support.
We also decided to multiply the value attributed to each treatment (for a given stage) by the standardized value of each action's action group, computed as follows:

$$\left[Q(s, a_g) - mean_{a_g}(Q(s, a_g))\right] / max_{a_g}\left[Q(s, a_g) - mean_{a_g}(Q(s, a_g))\right]$$

For brevity, let $s$ represent the entire state (i.e., the SWAT stage of the patient and the number of weeks remaining in their rehabilitation). $a_g$ represents an action group. The standardized value was used to aid in the interpretation of the results when using $\pi_{Mixed}$ (see below) but is not necessary for $\pi_{Agent}$ itself. If a group's state-action value is larger than another group's state-action value, its standardized value will still be larger. Thus, the rankings of the treatments (and the treatment selections) are unaffected by the standardizing process. We called the final value associated with each treatment in a given stage the standardized state-action value contribution (SSAVC) and selected treatments according to this value.

3. $\pi_{Mixed}$: Treatments were selected through a combination of the first and second policy definitions, represented by the following expression:
$$perceived\ treatment\ benefit + weight * SSAVC,$$
where $weight$ is a parameter used to alter the value placed on the agent's opinion. We considered integer weights from one to 20. The SSAVC was computed as described above so that it was easier to interpret the expression used here and the associated result. SSAVC is bounded above by one, which means that the adjustment to the physiotherapist's perceived treatment benefit is bounded by the weight.
4. $\pi_{Optimal}$: Treatments were selected according to their actual treatment benefit, which always results in the highest probability of transitioning to the next stage and hence maximized the return. In practice, choosing treatments in this manner is not possible, but using this process facilitated comparing our other treatment selection processes to the optimal process.

For $\pi_{Agent}$, the agents were trained using 1000 patients of training data. Since Parkwood Institute had collected data for approximately 90 patients in the few years prior to this study, we considered 1000 patients a reasonable upper bound on the number of patients from which treatment data can be collected before we require that the agent is able to learn something meaningful.

However, since the process of obtaining training data is ongoing, it is also of interest to see the impact of the size of the training dataset on the usefulness of the treatment selection processes. This is of particular interest for $\pi_{Mixed}$, which is how we expect a decision support system would be used in practice. Changing the size of the training set using this policy definition facilitates examining the change in the weight that should be given to the agent as the amount of training data increases. We used training set sizes of 100 patients up to 1000, using increments of 100. When increasing the size of the training set, we did not create an entirely new dataset; instead, we added 100 patients to the previous training set. To assess the variability of these treatment selection processes, we ran this entire simulation process 100 times using SHARCNET[1], changing only the training data each time.

## 4. Results

Figure 2 shows violin plots of the average return obtained from treating 1000 patients, using $\pi_{Agent}$ with the DKBG agent and the TEBG agent. The black dots represent the overall average return across all 100 simulation runs (i.e., an average of the average returns from treating 1000 patients). For both agents, the distributions were approximately symmetric, so their medians were very similar to their means. On average, the DKBG agent outperformed the simulated physiotherapists and the TEBG agent performed only slightly worse than the simulated physiotherapists.

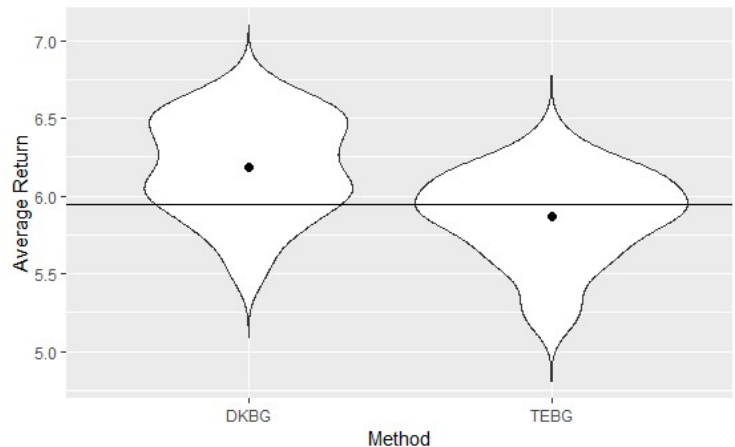

Figure 2: Violin plots of the average return obtained from treating 1000 patients, using $\pi_{Agent}$ with the domain knowledge based grouping (DKBG) agent and the treatment embedding based

---

[1] This research was enabled in part by support provided by Compute Ontario (www.computeontario.ca) and Compute Canada (www.computecanada.ca).

grouping (TEBG) agent. The horizontal line represents the average return under $\pi_{PT}$ and the black dots represent the overall average return across all 100 simulation runs (i.e., an average of the average returns from treating 1000 patients).

Since the simulator was designed such that a patient can only improve by one stage at a time, the transition probabilities provide another way of assessing the effectiveness of the methods. For each of the 100 simulations, the transition probability in each stage was calculated for the DKBG and TEBG agents' treatment selections. Table 3 shows the average of these transition probabilities for each stage[2]. Except for Stage Zero and Stage 10, the DKBG agent outperformed the TEBG agent. It is possible that the grouping procedure used in the TEBG approach performs particularly well in the edge stages, but further investigation is needed to determine if this is the cause of the TEBG agent's relatively strong performance for these stages. The performance of both agents deteriorated as the stage increases, which is an intuitive result because there were less training data for these stages. For comparison, the mean transition probability for each stage under $\pi_{PT}$ is also shown. Note that the physiotherapist treatment selections were designed such that the mean transition probability under $\pi_{PT}$ was 0.253 for every stage, so the observed differences are due to randomness[3]. The agents both outperformed the physiotherapists in general for the earlier stages but performed worse in the later stages.

| Stage | Mean Transition Probability | | |
|---|---|---|---|
| | Physiotherapist | DKBG | TEBG |
| 0 | 0.286 | 0.266 | 0.340 |
| 1 | 0.220 | 0.326 | 0.261 |
| 2 | 0.264 | 0.377 | 0.375 |
| 3 | 0.244 | 0.380 | 0.265 |
| 4 | 0.246 | 0.290 | 0.252 |
| 5 | 0.278 | 0.300 | 0.216 |
| 6 | 0.121 | 0.201 | 0.148 |
| 7 | 0.341 | 0.244 | 0.145 |
| 8 | 0.336 | 0.233 | 0.231 |
| 9 | 0.183 | 0.164 | 0.118 |
| 10 | 0.203 | 0.121 | 0.160 |

Table 3: Mean transition probabilities for each stage under $\pi_{PT}$ (physiotherapist) and $\pi_{Agent}$, using the domain knowledge based grouping (DKBG) and treatment embedding based grouping (TEBG) agents.

---

[2] To compute these probabilities, action selections for each stage must be determined. To obtain the action selections, input data, including the remaining number of weeks of treatment, must be given to the agents. We chose to use the number of weeks of treatment that corresponds to a patient that just began their treatment. It should be noted, however, that the number of weeks remaining has no impact on the actions selected, and thus no impact on the transition probabilities.

[3] We found it odd that the mean transition probability under physiotherapist treatment was so much more volatile for stages six through 10. Upon investigation, we found that this happened just due to chance. The stages with very low (high) transition probabilities had a best treatment (i.e., the treatment with the highest treatment benefit) with a relatively low (high) treatment benefit.

The two surface plots shown in Figure 3 illustrate the overall average return for $\pi_{Mixed}$. The left and right surfaces were created using the DKBG and TEBG agents, respectively. These surfaces show the relationship between the weight given to the agent, the number of training episodes, and the average return. The values of these features have been standardized, so only the shape of the surfaces should be considered (i.e., individual points on the left and right surfaces are incomparable). To illustrate this point, at 100 training episodes and a weight of one, the visualizations make it look like using the DKBG agent was inferior to using the TEBG agent, but this was not the case; their average returns were 6.034 and 6.023, respectively.

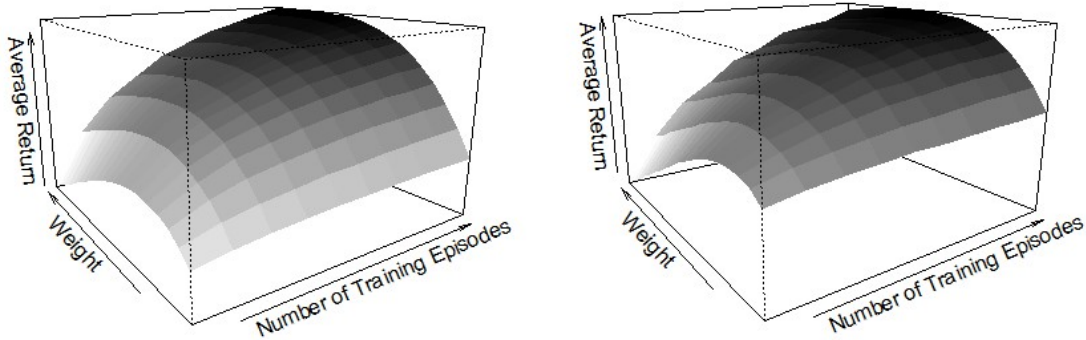

Figure 3: Surface plots of the overall average return obtained under $\pi_{Mixed}$ with the domain knowledge based grouping (DKBG) (left) agent and treatment embedding based grouping (TEBG) (right) agent as a function of the number of training episodes used for training and the weight given to the agent. Note that these plots are standardized so they cannot be compared to one another. In both, performance increases with number of training episodes, but the optimal return is achieved by selecting actions using a weighted combination of state action values and prior knowledge about anticipated treatment benefit.

For relatively few training episodes, both surfaces have an inverted U-shape, where medium-sized weights perform the best. At 100 training episodes and a weight of 20, using the DKBG and TEBG agents led to average returns of 5.796 and 5.538, respectively; both agents negatively impacted performance in this case, as the physiotherapists alone achieved an average return of 5.949. In at least one case, the fit of the regression model was rank-deficient. Although this is not particularly troubling when we are interested in prediction (as opposed to inference), it does indicate that the amount of data was insufficient. As the number of training episodes increased, both the optimal weight given to the agents and the overall average return increased as well. The maximum average returns using the DKBG and TEBG agent were 6.876 and 6.608, respectively, achieved with weights of 15 and 12, respectively. This indicates that the inverted U-shape shown for relatively few training episodes is still present with 1000 training episodes, albeit less clearly. In both cases, the maximum overall average return obtained was much larger than the overall average return the agents achieved on their own, shown in Figure 2.

In Figure 4, a cross-section from each of the two surfaces in Figure 3 is shown. Unlike in Figure 3, the two selection processes, one using the DKBG agent and the other using the TEBG agent, can be directly compared. The cross-section shown is the part of the surfaces where the weight is

11. This weight was chosen to strike a balance between reasonable effectiveness with relatively few training episodes and maximizing the use of the agent with relatively many training episodes. The darker shaded areas represent 95% confidence intervals and the lighter shaded areas are the areas within the fifth and 95$^{th}$ quantiles of the simulation results. For reference, the average return achieved under $\pi_{PT}$ and $\pi_{Optimal}$ are shown. Clearly, the process using the DKBG agent performed better than the approach using the TEBG agent, although both can be used to improve treatment selection. After 1000 training episodes, the DKBG and TEBG approaches reached overall average returns of 6.853 and 6.608, respectively, which corresponds to eliminating 54.1% and 39.4% of the gap between optimal treatment selection and physiotherapist treatment selection.

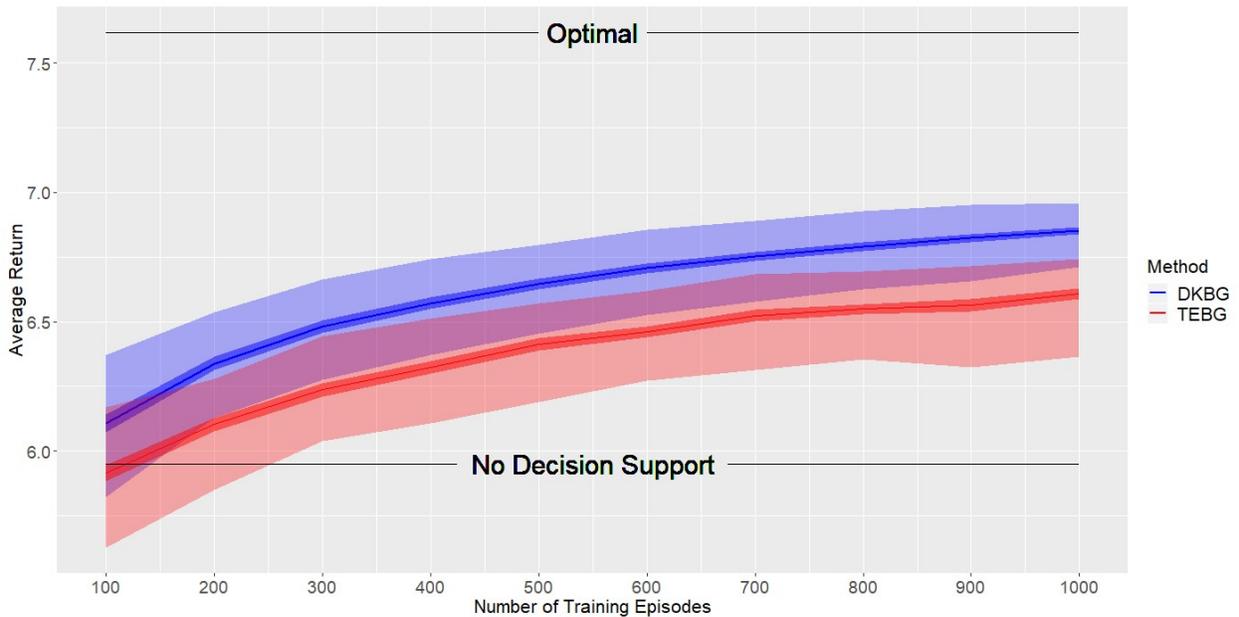

Figure 4: A plot of the overall average return under $\pi_{Mixed}$ (with a weight of 11) versus the number of episodes used to train the agent for both the domain knowledge based grouping (DKBG) and treatment embedding based grouping (TEBG) agents. The darker shaded areas represent 95% confidence intervals and the lighter shaded areas are the areas within the fifth and 95$^{th}$ quantiles of the simulation results. The horizontal lines show the average return under $\pi_{PT}$ (no decision support) and $\pi_{Optimal}$.

## 5. Discussion

The very large action space associated with SCI rehabilitation poses a challenge for the effective training of an RL agent, especially given the limited training data available. As shown in Appendix 2, training an agent that considers each action independently is ineffective, as the variance of the state-action value estimates are too high – resulting in optimistic state-action value estimates – even after 1000 training episodes. However, the results shown in Section 4 suggest that it is possible for an RL agent to learn something meaningful in this domain by grouping the actions. The results indicate that both agents we trained using grouped data, the

DKBG agent and the TEBG agent, can be used to augment SCI treatment selection relative to when physiotherapists choose treatments independently. The DKBG agent clearly outperformed the TEBG agent on average in all cases. Thus, if possible, using domain knowledge to group the actions seems preferable to grouping the actions using representation learning. If this is not practical, the TEBG agent can still be used to improve treatment selection, albeit to a lesser degree.

For both agents, using the agent to independently select treatments ($\pi_{Agent}$) was clearly less effective than combining the agent with the domain knowledge of the simulated physiotherapists ($\pi_{Mixed}$). We expect that the system would be used operationally in this way (i.e., a combination of the knowledge of the physiotherapists and the agent, albeit in a less explicit, numerical format), both because it has been shown to be the most effective in this work and because individual patients will have specific needs or limitations that are not incorporated into their state. For example, if a patient has additional trauma to their lower limbs, a standing exercise may not be possible, even though it may be the optimal treatment for this patient given the state of their SCI. The physiotherapist would know of this patient's limitation, but in its current state, the RL agent would not.

RL methods are typically used in situations where some actions may be beneficial in the long term, but not the short term. For example, a situation could arise where choosing action $a$ results in an immediate reward (and no further future rewards), but choosing action $b$ results in a larger, delayed reward (through transitioning to a new state where this reward is obtainable). With a discount factor of one, action $b$ is preferable even though it does not yield an immediate reward. RL approaches are designed to account for this delayed reward to maximize the long-term return. However, in our case, the simulator is structured in such a way that maximizing the long-term return is achieved simply by selecting the treatments that maximize the probability of transitioning to the next stage; actions cannot contribute to a good long term outcome without causing a good short term outcome. We assume that in SCI rehabilitation this is actually not the case, but using this simplified framework facilitates evaluating the methods on a stage-by-stage basis, as shown in Table 3. Although the simulator is set up in such a way that actions that contribute to a good long term outcome also cause a good short term outcome, the RL techniques used to learn from the data do not make use of this information. Thus, setting up the simulator in this way facilitates a useful way of assessing the techniques without giving them an unrealistic advantage.

The RL techniques do not have a known unrealistic advantage in the simulation relative to practice, but the improvement in patient outcomes from using decision support may be different in practice from what we observed in the simulation. Since the dynamics of optimal SCI rehabilitation are unknown, it is not possible to set all the simulator's parameters in a way that is guaranteed to accurately reflect SCI rehabilitation in practice. The true actual treatment benefits of each treatment are unknown, as is the correlation between the physiotherapists' perceived treatment benefits and the actual treatment benefits, and the transition function. The simulator also currently does not include interactions between treatments or delayed treatment effects, both of which we believe to exist in practice. Finally, conditioned on the actual treatment benefit, the physiotherapists' ranking of treatments are independent of each other. It is not clear how correlated physiotherapists' treatment rankings should be; there are different ideologies about SCI rehabilitation and rehabilitation programs are known to be heterogeneous, but the heterogeneity in treatment choices observed in our simulation is likely much higher than would be observed in practice. This would impact the data the agent has to learn from and could also impact the benefit of using the agent. Due to the independence of the treatment rankings, along with each physiotherapist having useful knowledge, a method whereby the most common treatments for each stage are chosen leads to excellent patient outcomes in our simulation, better than when using the agents (even in conjunction with the physiotherapists) to select treatments. However, with correlated treatment choices, this approach will be far less effective; understanding this relationship is an important avenue for future work.

All these issues impact the improvement in patient outcomes that we might see in practice. We plan to address these issues in future work, including investigating in which contexts the agents (in conjunction with physiotherapists) outperform the method of simply choosing the most common treatments selected by physiotherapists. However, due to these limitations, the magnitude of the improvement expected in practice cannot be estimated using the simulation results, and it is even possible that the methods we have developed would not be helpful after 1000 patients of training data. However, we showed that the simulated environment was too challenging for multiple alternative approaches, demonstrating that the simulated environment does not lead to an easy RL problem, and showed that the methods, when used in conjunction with physiotherapists, can provide value even after learning from only 100 patients. This provides reason to believe the methods may be useful in practice and motivates continued data collection in the field.

Decision support for SCI rehabilitation can be implemented in practice in multiple ways. Not only can it help physiotherapists in their day-to-day decision making, but it can also be used as a form of training students or new staff and to facilitate discussion among physiotherapists regarding their treatment practices. Establishing an environment that enhances a more reflective approach to practice and encourages constant improvement is a much-desired outcome across the field of physiotherapy. It could also provide guidance to outpatients, who continue their rehabilitation but do not see physiotherapists as frequently as inpatients.

One challenge with implementing decision support in practice will be balancing the agent's opinion with a physiotherapist's. It is clear from our study that the relative weighting is important, but it is not clear how to optimally choose that weighting. Another interesting aspect to consider for use in practice is balancing the exploration-exploitation trade-off. Although the agent will already be trained using the original dataset generated using only the physiotherapists' treatment selections, the agent can (and should be) continually trained as more data becomes available. There may be treatments that physiotherapists rarely use under a set of circumstances but are actually useful. Without exploring these treatments, we will never discover their value. The exploration aspect of this problem is unique due to the selection of multiple actions in each time step. The probability of transitioning to the next stage is dependent on all the actions selected in each week; thus, for an action $a$ in stage $s$, the estimated state-action value, $Q(s, a)$, is dependent on the actions selected with $a$ while in $s$. For this reason, the exploration component may include selecting a subset of actions while consciously choosing not to use another subset of actions. In addition, treatments must be selected in a manner that does not jeopardize the rehabilitation of the patient. Adopting a Bayesian RL approach might help with these problems. It provides a natural way to both incorporate prior knowledge and optimize the exploration-exploitation trade-off. An added benefit is that incorporating prior knowledge in this fashion might also allow the agent to learn more efficiently.

The initial findings associated with this work have substantial implications for guiding our future work. Importantly, although the simulation methods necessarily required several assumptions and also resulted in identifying specific challenges, we are confident that this approach has set us off on an exciting and worthwhile journey that will have considerable impact for the field of neurological rehabilitation. In many ways, this reflects a "holy grail" of this field in unravelling the "black box" of rehabilitative therapeutic approaches through the development of a learning system focused on the identification of the combination of therapies that would be associated with achieving the best possible patient outcomes. The potential applications for this are significant. For the physiotherapist, such a system would assist clinical decision-making and reflective practice informing a personalized treatment approach that would be part of a learning health system continuously optimizing and improving practice to the best possible outcome. Some key benefits include identification of appropriate time points for re-assessment, flagging anomalies which should be further investigated through medical review, and identifying what may inhibit recovery. This would also have significant utility in training new staff and trainees. For the patient, we see the potential for more active involvement in the rehabilitation process through goal identification and continuous monitoring towards desired outcomes. This represents a novel form of evidence generation that would support knowledge translation and adherence to best practice.

Although significant work lies ahead to achieve these lofty ambitions, it is clear that important next steps will be to bring together data scientists, physiotherapists, and persons with lived experience of SCI, possibly within a Bayesian approach, to further refine or extend the model – especially given the current finding that domain-specific knowledge appears to be useful for these refinements. Currently, we are embarking on a multi-site investigation to bring individuals together who represent several specialized rehabilitation programs to inform continued evolution of this work. Specific issues that will be addressed will include further characterization of specific combinations of therapeutic approaches and their relative merits (i.e., weighting) as well as strategies for standardizing and enhancing feasibility of data capture across different sites that provide SCI rehabilitation, which will enable access to larger datasets. Future work may also examine such methods as incorporating wearable technologies to automate data capture as well as to better understand the implications of activities outside of the therapeutic environment. A key component of this will be incorporating patient-driven goals through the automated integration or patient-reported outcome measures (PROMs) as part of these efforts.

## 6. Conclusion

Applying RL to treatment selection for SCI rehabilitation is challenging because limited data are available, there are many possible treatments to choose from, and treatments are chosen simultaneously. We have demonstrated that some more traditional approaches using Fitted Q Iteration are not able to effectively learn in this environment. However, treatments for SCI have similarities that allow them to be grouped. We have proposed two methods for grouping the treatments, DKBG and TEBG, that facilitate learning about a treatment even if it was not chosen. We then used Fitted Q Iteration to train agents after this grouping procedure. When the agents were used in conjunction with the simulated physiotherapists, they were able to contribute to improved patient outcomes in our simulated environment relative to when the physiotherapists acted alone, with the DKBG agent offering larger improvement than the TEBG agent. This is an exciting first step in the application of RL to this field, and there are many possible avenues of continued research that will require collaboration between physiotherapists and other clinicians, persons with lived experience (i.e., patients), and computer scientists.

# Appendix 1

**Simulator for physical rehabilitation of spinal cord injury**

In the case of spinal cord injury (SCI) rehabilitation, the environment is the patient. The states reflect the current health status as relevant for treatment and the actions correspond to treatment choices. The state that each patient is in is defined by two components, their stage and the number of weeks remaining in their rehabilitation. There are 12 stages ranging from zero, the most impaired, to 11, a terminal state indicating unimpaired mobility (i.e., no standing or walking impairment). These stages were chosen to reflect the 12 stages of the SCI Standing and Walking Assessment Tool (SWAT) [7], [8]. Each patient has a limited number of weeks of treatment. Based on their initial stage, they are afforded a maximum number of treatment weeks according to the following formula:

$$maximum\ number\ of\ weeks = floor[1.25(11 - stage)] \qquad (2)$$

This formula was chosen to reflect the grouping methodologies used by rehabilitation hospitals and similar health care facilities and to facilitate a mean length of inpatient stay that approximates the mean stay in the data obtained from Parkwood Institute. After each week of treatment, a patient's remaining number of weeks is decremented by one. It should be noted that the formula we use in the simulator is a simplification of the actual methodologies used in practice; however, it could easily be substituted with the formula used in a particular health care system/setting.

In clinical practice, there are a myriad of potential treatments (from tens to hundreds depending on how treatments are classified), but they can be grouped in a meaningful way in terms of their appropriateness for patients with different health statuses. In the simulator, we chose to have 110 treatments numbered 1 to 110, each placed into one of 11 groups. The first group is composed of treatments 1 to 10, the second group is composed of treatments 11 to 20, and so on. For each stage each group is assigned a ranking (lower is better), indicating how useful treatments from these groups are expected to be if applied to a patient in that stage. As an example, for Stage Four patients, treatments 41 to 50 (the fifth group) are given a ranking of one, treatments 31 to 40 and 51 to 60 (groups four and six, respectively), are given a ranking of two, and the remaining treatments are given a ranking of three. Table A1 shows the ranking of each treatment group for Stage Four. Using this approach, each stage has 10 actions expected to be the best, 20 actions expected to be second to these top 10 actions, and another 80 actions expected to be inferior to the top 30. To maintain this distribution of actions for the edge stages (zero and 10, since 11 is terminal), the two groups nearest the group ranked first are both given a ranking of two.

| Group   | 1 | 2 | 3 | 4 | 5 | 6 | 7 | 8 | 9 | 10 | 11 |
|---------|---|---|---|---|---|---|---|---|---|----|----|
| Ranking | 3 | 3 | 3 | 2 | 1 | 2 | 3 | 3 | 3 | 3  | 3  |

Table A1: This table outlines the ranking (lower is better) of each treatment group for a patient in Stage Four.

To simulate the effectiveness of treatments, it is necessary to have some method for differentiating one treatment from another. It is unreasonable to assume that every treatment from each group is equally useful for each stage. For this reason, we use the idea of a *treatment benefit* to provide a numerical value indicating the usefulness of each treatment in each stage. In practice, these treatment benefits are unobservable, so they are used only to simulate experience and are not involved in the process of learning an improved policy.

Each treatment has 11 associated treatment benefits, one for each non-terminal stage. The treatment benefits are generated from a normal distribution with mean and standard deviation set based on the treatment's ranking. Table A2 shows the mean and standard deviation associated with each treatment ranking. The standard deviations increase with treatment ranking because we assume that treatment benefits vary more for treatments that are not specifically intended for the given stage. The actual treatment benefits are randomly generated once, at the beginning of the simulation, and are constant for the entire simulation.

| Treatment Ranking | Mean | Standard Deviation |
|---|---|---|
| 1 | 7.0 | 0.75 |
| 2 | 5.5 | 1.25 |
| 3 | 4.0 | 1.50 |

Table A2: The actual treatment benefits are simulated from a normal distribution using the parameters associated with their treatment ranking. This table shows the parameters associated with each rank.

To simulate the presence of domain knowledge about different treatments, we introduce the idea of a perceived treatment benefit. Perceived treatment benefits represent a physiotherapist's opinion on the benefit of each treatment for each stage. Since physiotherapists are knowledgeable about these treatments, the perceived treatment benefits are designed to be correlated with the actual treatment benefits. They are generated from a conditional normal distribution, conditioned on the actual treatment benefits. The correlation and unconditional mean and standard deviation for each treatment ranking are shown in Table A3. With $\rho$ representing the correlation and $ATB$ representing the actual treatment benefit, the conditional mean and standard deviation are computed as follows:

$$\begin{aligned} Conditional\ Mean \\ = Unconditional\ Mean \\ + \rho(Unconditional\ Standard\ Deviation) \\ /(Standard\ Deviation_{ATB})(ATB - Mean_{ATB}) \end{aligned} \quad (3)$$

$$\begin{aligned} Conditional\ Standard\ Deviation \\ = \sqrt{(1-\rho^2)}(Unconditional\ Standard\ Deviation) \end{aligned} \quad (4)$$

Like with the actual treatment benefits, the perceived treatment benefits vary more for treatments that are not specifically intended for the given stage. Unlike the actual treatment benefits, perceived treatment benefits are generated for each iteration of the simulation, representing differences in individual physiotherapists' opinions.

| Treatment Ranking | Correlation | Unconditional Mean | Unconditional Standard Deviation |
|---|---|---|---|
| 1 | 0.8 | 7.0 | 1.00 |
| 2 | 0.7 | 5.5 | 1.50 |
| 3 | 0.5 | 4.0 | 1.75 |

Table A3: The perceived treatment benefits are simulated from a conditional normal distribution based on the actual treatment benefits. The correlation between the perceived treatment benefits and the actual treatment benefits and the unconditional mean and standard deviation for each treatment ranking are shown in this table.

The scatter plot in Figure A1 shows the actual treatment benefits in Stage Zero and the associated perceived treatment benefits that a physiotherapist might have. This plot clearly shows that the treatments with a larger rank exhibit a much weaker correlation between actual and perceived treatment benefits.

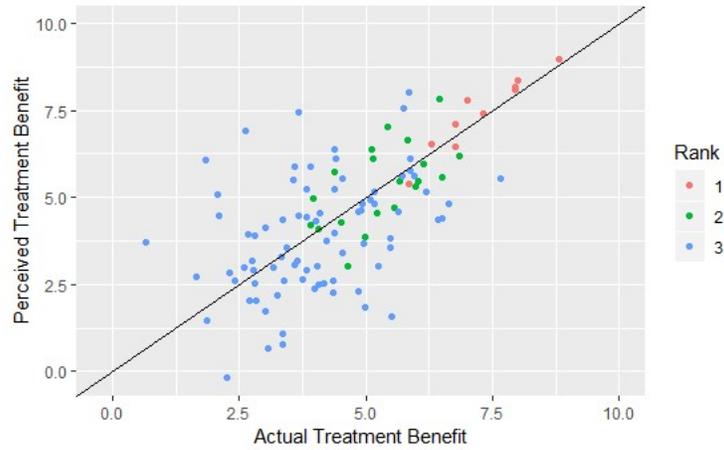

Figure A1: A scatter plot of the perceived treatment benefit versus actual treatment benefit for a patient in Stage Zero, categorized by their treatment ranking.

For a given treatment period, a physiotherapist selects multiple treatments to use. The aggregate treatment benefit refers to the sum of the individual treatment benefits of the selected treatments.

From a stage $x$, there are only two possible next stages, $x$ and $x+1$. The probability of a transition from $x$ to $x+1$ is conditioned only on the aggregate treatment benefit of the selected treatments. The probability, $p$, of a transition to the next stage is computed using the formula shown below and is visualized in Figure A2:

$$p = \frac{2}{3}\left[\frac{1}{1 + e^{-(0.2339304 - 13.49396(aggregate\ treatment\ benefit))}}\right] \quad (5)$$

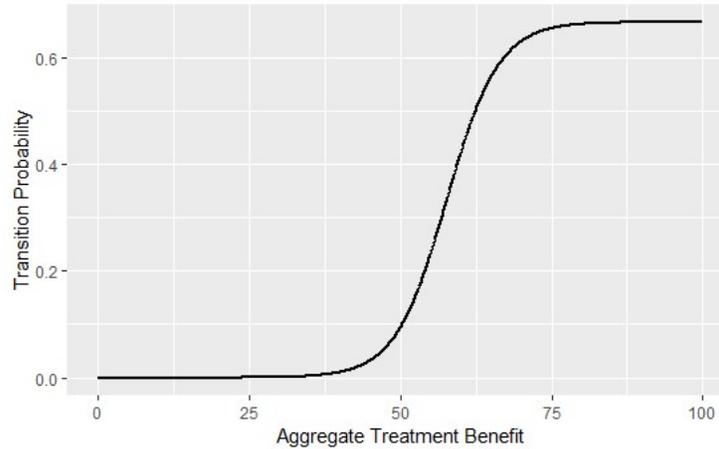

Figure A2: The probability of a patient's transition to the next stage as a function of their aggregate treatment benefit.

This transition function is designed to meet three criteria:

- In accordance with the data received from Parkwood Institute, the mean probability of transition using the physiotherapists' treatment selections should be in the range of 0.25-0.30. Note that this range is a coarse estimate, as it is based on the cumulative improvement in stage from admission to discharge for all patients divided by the cumulative number of weeks of treatment. Under simulated physiotherapist treatment selection, the mean transition probability is approximately 0.253.
- The transition probability should be very small, but non-zero, when treatments are selected uniformly randomly. Uniform random treatment selection results in a mean transition probability of approximately 0.008.
- Regardless of the treatments chosen, the transition probability should never approach 1 because the body requires time to heal. With this transition function, the transition probability cannot exceed 0.667.

For the first two criteria, it must be noted that transition probabilities are dependent on the set of actual treatment benefits and that these mean transition probabilities have been computed across these sets. Thus, the mean transition probabilities under a single realization of a set of actual treatment benefits (which is what is used in this study) are not necessarily the values stated.

The experience data is generated using episodes. An episode is a patient's entire treatment period. An episode begins with the random selection of a non-terminal stage. Based on the data obtained from Parkwood Institute, the initial stage is sampled from a distribution where Stage Zero is selected with a probability of $\frac{2}{7}$. The remaining ten non-terminal stages each account for $\frac{1}{14}$ of the distribution's probability mass.

A set of perceived treatment benefits is generated for the given stage. From this, a treatment plan is created, representing a course of treatment that a physiotherapist might plan for a week. The treatment plan is composed of the eight treatments with the highest perceived treatment benefit.

An episode terminates when a patient either reaches the final (fully healthy) stage or after their allocated number of weeks has elapsed. Figure A3 shows six possible patient trajectories through state space. An episode ends upon reaching either the horizontal (out of time) or vertical (unimpaired mobility) dashed line.

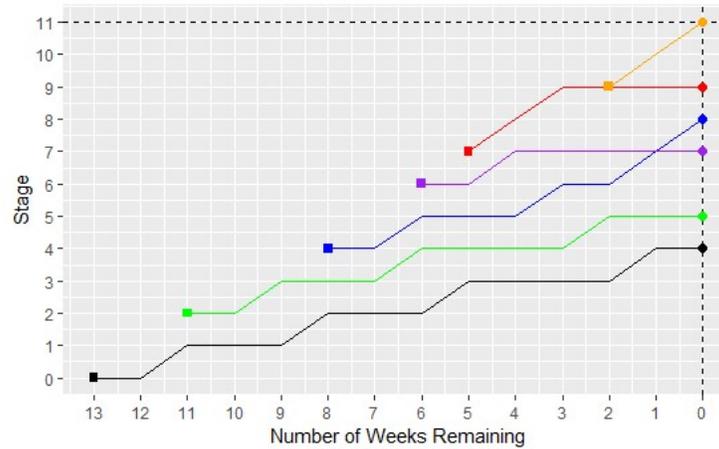

Figure A3: Possible trajectories through the state space for six simulated patients. The start and end of the episodes are represented by a square and circle, respectively.

# Appendix 2

**Alternative reinforcement learning approaches**

We considered multiple alternative approaches to the ones presented in the main body of the paper. These were considered with 1000 patients of training data.

Using Fitted Q Iteration (see Section 3.2.5 for a description) to learn an optimal policy for treatment selection involves fitting a regression model to learn the state-action values. Perhaps the most natural choice of model is the following, with $\boldsymbol{a}$ denoting a vector representing the selected actions, $\mathcal{A}$ denoting the entire action space, and $\mathcal{S}$ denoting all stages:

$$
\begin{aligned}
Q(s, \boldsymbol{a}, & No.\,of\,Weeks\,Remaining) \\
&= \beta_0 + \beta_1(No.\,of\,Weeks\,Remaining) \\
&\quad + \sum_{i=1}^{8}\left(\sum_{action\,\in\,\mathcal{A}\backslash 1} \beta_{action} I(action = \boldsymbol{a}_i)\right. \\
&\quad + \sum_{stage\,\in\,\mathcal{S}\backslash 0} \beta_{stage} I(stage = s) \\
&\quad + \left.\sum_{action\,\in\,\mathcal{A}\backslash 1,\ stage\,\in\,\mathcal{S}\backslash 0} \beta_{action,\,stage} I(action = \boldsymbol{a}_i) I(stage = s)\right)
\end{aligned}
\tag{6}
$$

Note that the first stage and action are omitted because we used dummy coding here. This model has two problems. With our simulation settings, the model has 1211 coefficients, which makes it rank-deficient with the limited data available. Rank-deficiency does not mean we are unable to use the model to make predictions, but it does suggest that the amount of data is insufficient for the model. The second, bigger problem is that the Fitted Q Iteration algorithm diverges with this regression model. This is a well-known issue with Fitted Q Iteration when using a linear model (e.g., [34]). One approach to solving the first problem is to group the treatments using the approaches described in Section 3.2.1 and Section 3.2.2, as this reduces the number of coefficients needed to fit the model. With groups formed using either process, this model would look as follows, with $\boldsymbol{a}_g$ denoting a vector representing the selected action groups, $\mathcal{G}$ denoting all action groups, and $\mathcal{S}$ denoting all stages:

$$\begin{aligned}
Q(s, &\boldsymbol{a}_g, No.\,of\,Weeks\,Remaining) \\
&= \beta_0 + \beta_1(No.\,of\,Weeks\,Remaining) \\
&\quad + \sum_{i=1}^{8}\left(\sum_{actionGroup\,\in\,\mathcal{G}\backslash 1} \beta_{actionGroup} I(actionGroup = \boldsymbol{a}_{g_i})\right. \\
&\quad + \sum_{stage\,\in\,\mathcal{S}\backslash 0} \beta_{stage} I(stage = s) \\
&\quad + \sum_{actionGroup\,\in\,\mathcal{G}\backslash 1,\ stage\,\in\,\mathcal{S}\backslash 0} \beta_{actionGroup,\ stage} I(actionGroup \\
&\quad \left.= \boldsymbol{a}_{g_i}) I(stage = s)\right)
\end{aligned} \qquad (7)$$

Like in the above equation, we used dummy coding for the action groups, so the first action group is omitted. We omit terms related to the first action group because the number of actions always sums to eight for our data. With grouping, the model is no longer rank-deficient but the Fitted Q Iteration algorithm still diverges using this model.

Another option is to use the model described by Equation 6 with the data splitting approach described in Section 3.2.3. With this setup, the regression model is rank-deficient but Fitted Q Iteration converges. The state-action value estimates we obtained are very high, suggesting huge improvements from choosing treatments selected by the agent instead of the physiotherapists. However, the state-action value estimates are overly optimistic, as the treatments this agent recommends actually give a patient very little chance of making progress in their recovery. Our rationale for this observation is that the estimates of the optimal state-action value function for Stage 10, the last non-terminal stage, are highly variable because some actions were selected very few times at this stage, even after 1000 episodes. Consequently, some state-action value estimates for actions in Stage 10 are very high. In turn, this results in other actions also having an optimistic state-action value estimate in Stage 10 because the agent believes that, even if the patient does not reach unimpaired mobility immediately, it will be able to select actions next week that will help the patient reach unimpaired mobility. This optimism percolates to the state-action value estimates in Stage Nine, then Stage Eight, and so on, all the way down to Stage Zero. As a result, the agent trained in this manner is not useful.

# Appendix 3

## Data adjustment: Separating the actions

We adjusted the dataset by mapping each row of data to eight rows, one for each action in the original row. To test the efficacy of this approach, we developed a simple case with only three actions and two stages. Two actions are chosen in the first stage, and the probability of moving to the next stage depends on the actions according to the values in Table A4.

| Number of Selections | | | Probability of Success |
| --- | --- | --- | --- |
| Best Action | Middle Action | Worst Action | |
| 2 | 0 | 0 | 0.8 |
| 1 | 1 | 0 | 0.7 |
| 1 | 0 | 1 | 0.55 |
| 0 | 2 | 0 | 0.6 |
| 0 | 1 | 1 | 0.45 |
| 0 | 0 | 2 | 0.3 |

Table A4: The probability of transition to the next stage as a function of the actions chosen.

In this situation, the Fitted Q Iteration algorithm converges even without separating the actions, providing a benchmark with which to compare our approach. With 1000 observations to learn from, both the benchmark approach and our approach obtain the same ranking of actions, with both correctly ranking them. Although this is a much simpler situation than spinal cord injury (SCI) rehabilitation and is by no means a proof that both approaches rank actions in the same order in all cases, it provides some reassurance that our approach ranks actions in a reasonable manner.